# Forward Collision Warning Systems – Validating Driving Simulator Results with Field Data

**Snehanshu Banerjee***
Department of Transportation & Infrastructure Studies
Morgan State University, Baltimore, MD 21251
ORCID ID- 0000-0002-2932-3618
Email: snban1@morgan.edu

**Mansoureh Jeihani**
Department of Transportation & Infrastructure Studies
Morgan State University, Baltimore, MD 21251
Email: mansoureh.jeihani@morgan.edu

* Corresponding Author

**Abstract**

With the advent of Advanced Driver Assistance Systems (ADAS), there is an increasing need to evaluate driver behavior while using such technology. In this unique study, a forward collision warning (FCW) system using connected vehicle technology, was introduced in a driving simulator environment, to evaluate driver braking behavior and then the results are validated using data from field tests. A total of 93 participants were recruited for this study, for which a virtual network of South Baltimore was created. A one sample t-test was conducted, and it was found that the mean reduction in speed of 15.07 mph post FCW, is statistically significant. A random forest, machine learning algorithm was found to be the best fit for ranking the most important variables in the dataset by order of importance. Field data obtained from the University of Michigan Transportation Research Institute (UMTRI), substantiated the FCW findings from this driving simulator study.

## 1. Introduction

Rear-end collisions are considered to be the most frequent and one of the deadliest type of road vehicle crashes as it accounted for almost half the two-vehicle crashes in the US between 2012 and 2014 [1], reaching more than 2.1 million crashes in 2015 and resulted in 6.8%, 32.4%, and 33.9% of the total two-vehicle crashes' fatalities, injuries, and property damage, respectively [2]. These high numbers of the severe consequences associated with rear-end collisions could be worse, if the other types of front-end crashes are involved; for example, head-on collisions accounted for 10.2% of fatalities in 2015 [2]. Several studies cited driver distraction or inability to react at the right time when the vehicle in the front brakes suddenly, as the two major reasons behind these high number of collisions [3, 4]. Hence, in order to reduce the number and/or severity of these crashes, ITS applications that alert drivers when they are too close to the vehicle ahead have been developed; these applications are called forward collision warning (FCW) systems [5]. These systems comprise of sensors installed at the front of the vehicle and used to scan the roads for vehicles that are close enough to warrant a danger of a crash and, consequently, issues a warning through visual, audio, and/or tactile means [3, 5]. Forward collision warning has been described by the National Highway Traffic Safety Administration as "one intended to passively assist the driver in avoiding or mitigating a rear-end collision via presentation of audible, visual, and/or haptic alerts, or any combination thereof" [6]. A forward collision warning system has the capability of detecting a vehicle in front using sensing technologies like LIDAR, radar, and cameras. After processing and analyzing sensor data, an alert is provided if there is a possibility of collision with another vehicle [7].

    According to Patra et al. [8], the earliest of these systems were based on radar technology to detect possible collisions which are in the line of sight of the vehicle. However, to overcome this problem, continuous advancement in these systems were achieved by first developing beacon-based FCW, then vision-based ones as well as cooperative collision warning systems, and lately, V2V systems. Hence, several research studies have been conducted in order to test the effectiveness of the FCW systems in enhancing the safety for the road users, through assessing the change in the drivers' performance and behavior while using these systems. Although there are some studies that tested and evaluated the performance of the system itself, the scope of this study

only focuses on the change in driver's behavior and performance measures using a driving simulator and its validation using field tests.

Driving simulator studies have been successfully used in prior researches to analyze driver behavior [9-15]. Most of the studies that attempted to assess the change in drivers' behavior and performance after the use of FCW were conducted using a driving simulator. The earliest studies conducted in this field were the ones done by Burns, Knabe and Tevell [16], and Lee, McGehee and Brown [17]. In the first study, the researchers found that drivers' who were presented with auditory and visual warnings traveled at slower speeds and had longer headways with the lead vehicle [16]; while, in the second study, the results showed that the FCW system had a significant effect on the time to release the accelerator by the driver which took, on average, 1.5s while using the system versus, approximately, 2.2s without the use of the system. Nevertheless, the FCW system had no significant effect on the initial accelerator release to initial brake press time, initial brake press to maximum braking time, and mean deceleration after initial brake press [17]. Moreover, Shinar and Schechtman [18] tested an FCW system that displayed a red warning light when the headway was 1.2s or less, and a buzzer sound whenever the headway was reduced to 0.8s. Through their driving simulator tests, it was found that the system had a significant impact on the drivers' headway as it reduced the time the drivers' spent at the shorter and more dangerous headway by 25% and increased the time they spent in the longer and safer one by 14%; and that these results did not differ according to either the gender or age of the drivers. Similar results were also obtained by Maltz and Shinar [19] as drivers' spent only 7% of the time in the short headway versus 12% without the use of the system, and they correctly slowed down for 86% of the instances of true alerts.

Few other studies tested the performance of a number of systems or different drivers' characteristics against each other. For instance, Lees and Lee [20] analyzed the driving performance in terms of speed, accelerator, and brake positions to three FCW systems that had different degrees of reliability: false alarms, unnecessary alarms, and accurate alarms. Through the simulator tests, it was found that the drivers who braked more frequently and reduced their speed more were the ones using the system with unnecessary alarms. Another study that tested different types of systems was the one conducted by Jamson, Lai and Carsten [21]. In this study, the researchers tested the reaction of the drivers to two different FCW systems: a non-adaptive FCW and an adaptive one that adjusted the timing of its alarms according to each driver's reaction time, through measuring their brake reaction time and minimum headway. The researchers found that the systems had no significant effect on the brake reaction time; yet, aggressive drivers had shorter reaction time with the adaptive system. On the other hand, the system had a significant impact on the minimum headway as those drivers who did not use the system had a headway of 1.11s versus 1.66s and 1.45s for the non-adaptive and adaptive FCW systems, respectively. Koustanai et al. [22] tested the drivers' speed, headway, and deceleration when they were familiarized with the system before using it versus when they were not. Through this study, the researchers found that, first, the use of FCW significantly impacted the brake time, headway and deceleration; while familiarized drivers maintained longer headway than unfamiliarized ones; approximately 5s versus 4.5s. Lastly, Muhrer, Reinprecht and Vollrath [23] developed an FCW with a braking intervention, called FCW+, and tested its performance, in terms of number of collisions, and its impact on the drivers' performance, in terms of reaction time, headway, and gaze behavior. Through the analysis of the results, it was found that the system had no significant impact on either the reaction time or headway, albeit it did not lead to more engagement in secondary tasks and led to significantly fewer collisions.

"Time-to-collision" (TTC), one of the most frequently used time-based surrogate safety measures, has been used consistently to develop and enhance road safety measures [24, 25]. Behbahani et al. [24] and Nadimi et al. [25] have used microsimulation techniques to analyze the usefulness of in-vehicle collision avoidance warning systems (IVCAWS) to reduce the risk of rear-end and angular collisions as well as modify existing TTC frameworks for increasing the precision and accuracy of detecting potential collisions.

Regarding the studies that were based on field tests, Ben-Yaacov, Maltz and Shinar [26] conducted a field experiment to measure the drivers' headway, speed, and reaction time when using an FCW system. The results of these experiments showed that the system had a significant impact on the time drivers spent in the danger headway zone, from 0 to 0.8s, as it dropped from 42.2% when not using the system to only 3.5% during the use of the system, were more likely to reduce their speed when the system send the warning, and that they slowed down more than they sped up during the use of the system when they were in the headway borderline zone, from 0.8s to 1.2s. Another field test study was conducted by the National Highway Traffic Safety Administration (NHTSA) as part of the Automotive Collision Avoidance System Field Operational Test (ACAS FOT) project [27]. In this comprehensive project, the ability of FCW to reduce the number of crashes was measured through its impact on the drivers' headway, approach behavior, braking behavior, and secondary task behavior. Regarding the former measure, it was found that the use of the FCW system led to a statistically significant increase in headway time as the average percentage of time spent while following a car at less than 1s headway dropped from 30% to 25.5% when the system was used. On the other hand, there was no statistically significant impacts of the system on the other three measures. Finally, Crump et al. [28] assessed the performance of an FCW system by testing the ability of the drivers to respond to the warning before an autonomous braking system is engaged. Through this field test, the system's effectiveness was proved as 14 out of the 16 drivers were able to react before the autonomous braking system is engaged with a median reaction time of 284 milliseconds; albeit seven drivers reacted, before the warning was sent. The FCW was programmed to occur in both the developed scenarios, where only the first stage of an FCW system was replicated for evaluation.

Thus far, the sample size of simulator-based studies related to FCW have been extremely limited, ranging from 20 – 64 participants and setting constraints on the participants on how they need to drive, such as driving at a certain speed [16, 18, 20-23]. This study takes a unique approach to analyze driver behavior in a driving simulator, when subject to an FCW system, by validating the simulator findings with the help of data collected from field tests, using a large sample size of participants and letting the participants drive as they normally would, without any constraints.

## 2. Methodology

A medium-fidelity full scale driving simulator at the Morgan State University Safety and Behavioral Analysis (SABA) Center was used for this study, to analyze driver behavior in response to an FCW system. The VR-Design studio software developed by FORUM8 Co. [29] was used to develop a virtual network of South Baltimore in Maryland. The idea behind choosing this location was such that, majority of the participants being familiar with the Baltimore area, the virtual network creates a more realistic driving experience for them. The simulator has the capability capture data such as steering wheel control, braking, acceleration and speed among others. The driving simulator is shown in Figure 1.

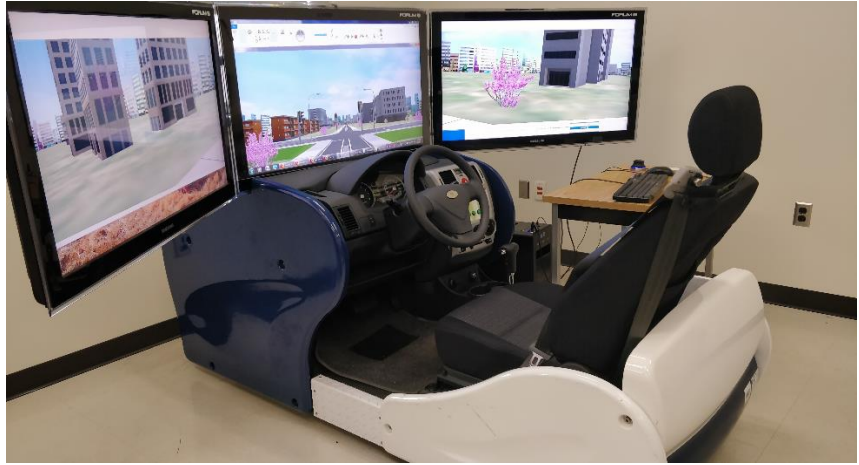

**Figure 1.** Driving simulator at Morgan State University and Tobii Pro eye tracking glasses

## 2.1. FCW

This application can utilize both Vehicle to Vehicle (V2V) or Automated Vehicle (AV) technology to warn the driver of an impending collision with a vehicle or object directly in its path. This study used the V2V feature of connected vehicle technology. The three stages of an FCW system are illustrated in Figure 2.

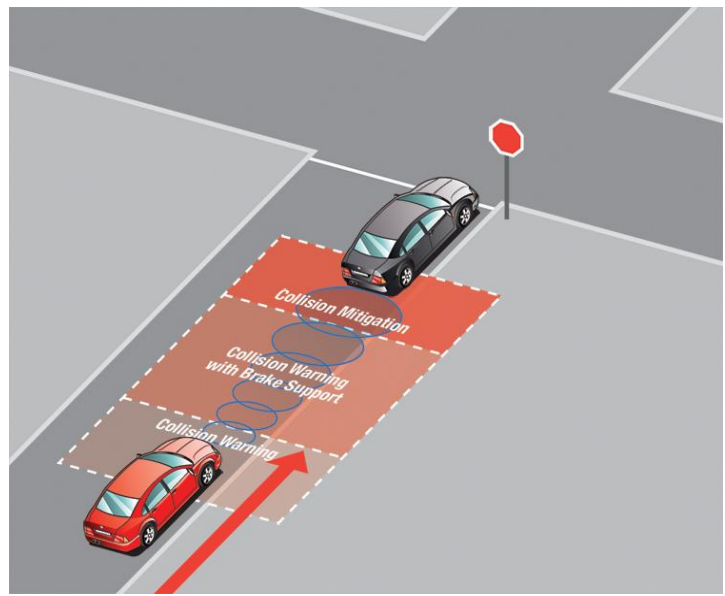

**Figure 2.** Forward Collision Warning

As shown in Figure 9, there are three stages in FCW: collision warning, collision warning with brake support, and collision mitigation. Due to the limitations of this driving simulator, only the first stage of an FCW system could be recreated for evaluation. This means that an FCW system, in this case, will not take any automatic action to avoid a collision or control the vehicle; therefore, post FCW, drivers will remain responsible for the safe operation of their vehicles to avoid a crash. The advantage of using a one-stage warning system is twofold: one, to warn a distracted driver and two, to maintain the driver trust in the system, which could be in jeopardy with the false alarm rates in the multi-stage system.

This event was programmed to occur in both the scenarios, as the probability of such an event occurring is totally dependent on the individual participants' driving behavior. Since the goal of evaluating this application was to analyze the influence of FCW on change in speed, a perception reaction distance as defined by the National Association of City Transportation Officials (NACTO) was used to identify the appropriate timepoints to send an FCW to the driver based on the speed of the vehicle. The perception reaction distances, as replicated in the driving simulator, were based on the respective speeds as shown in Table 1.

**Table 1.** Perception Reaction Distances

| MPH | Perception Reaction Distance (ft) |
|---|---|
| 10 | 22 |
| 15 | 33 |
| 20 | 44 |
| 25 | 55 |
| 30 | 66 |
| 35 | 77 |
| 40 | 88 |
| 45 | 99 |
| 50 | 110 |
| 55 | 121 |
| 60 | 132 |
| 65 | 143 |
| 70 | 154 |

Source: [30]

A snapshot of an FCW event is shown in Figure 3.

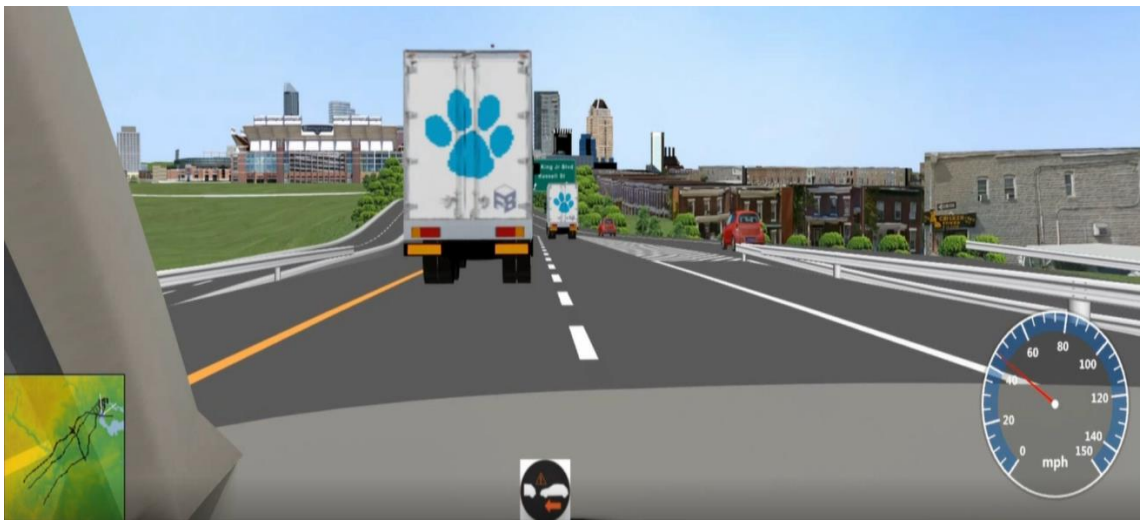

**Figure 3.** FCW snapshot in the driving simulation

Thus, the FCW in the driving simulator was activated, based on the perception reaction distances and the respective speeds, shown in Table 1.

## 2.2. Recruitment Process

Institutional Review Board (IRB) approval was received before participants were recruited for this study. Participants signed a consent form before participating in the study and were paid at the rate of $15 per hour of driving. A total of ninety-three participants from diverse socio-economic backgrounds took part in this study. Participants were recruited through a combination of emails to participants from prior studies [11, 12, 31-33] and distribution of flyers across the university and throughout Baltimore County. The details of the study were withheld so as to avoid driving bias, but they were given an opportunity to get familiar with the driving simulator.

## 2.3. Descriptive Statistics

This study involving 93 participants consisted of a balanced group of male and female individuals. Table 2 presents some of the sociodemographic statistics of the participants.

**Table 2. Participant socio-demographics**

| Variables | Characteristics | Percentage |
|---|---|---|
| Gender | Female | 44 |
|  | Male | 56 |
| Age | 18-25 | 37 |
|  | 26-35 | 29 |
|  | 36-45 | 14 |
|  | 46-55 | 12 |
|  | >55 | 8 |
| Education Level | High School or less | 12 |
|  | College degree | 61 |
|  | Post-graduate | 27 |
| Household income level | <$20,000 | 27 |
|  | $20,000 - $49,999 | 34 |
|  | $50,000 - $99,999 | 22 |
|  | >$100,000 | 17 |

## 2.4. Random Forest model

Random forest is a supervised learning algorithm which can be used for both classification and regression modeling [34]. This algorithm consists of an ensemble of decision trees, i.e., CART (classification and regression trees). It is commonly trained with the bagging technique in which the idea is to combine multiple models to improve classification accuracy, thereby reducing the risk of overfitting [35]. The decision trees in a random forest are trained on bootstrap sample sets produced from bagged samples. Once the set of decision trees has grown, the unsampled observations are dropped down each tree from the test dataset and these 'out of bag' (OOB) observations are used for internal cross validation and to calculate prediction error rates. The error calculated is the mean decrease in node impurity (mean decrease Gini or MDG) which can be used for variable selection by ranking variables in the order of importance. The random forest package in "R" [36] was used to compute MDG which is the sum of all decreases in Gini impurity due to a given variable and then normalized toward the end of the forest growing stage. MDG is the predictive accuracy lost by permuting a given predictor variable from the tree used to generate predictions about the class of observation $i$, where $i \in [0,1]$, the Gini score range. Thus, predictor variables with a higher MDG score more accurately predict the true class of observation $i$ which is also termed as the variable importance measure (VIM) in random forests.

## 3. Results

A total of 104 instances of FCW were detected in 186 experiments in which the participants were approaching the vehicle preceding them at an alarming speed. This data was extrapolated using a novel data extraction software [37]. In this analysis, average speeds were calculated 5 seconds before and after the FCW was issued, to evaluate the impact of the warning in terms of change in speed. The difference in speed change was considered in lieu of average before and after speeds, since the speed limits varied at different segments in the scenario and thus would not be a good measure for this analysis.

### 3.1. One sample t-test

A one sample t-test was conducted to determine whether the mean difference in speed change is statistically different from the hypothesized mean difference in speed of zero.

**Table 3.** One sample t-test

| | Hypothesized Mean Difference = 0 | | | | | |
|---|---|---|---|---|---|---|
| | | | | | 95% Confidence Interval of the Difference | |
| | t | df | Sig. (2-tailed) | Mean Difference | Lower | Upper |
| Change in speed | 12.990 | 103 | 0.000* | 15.070 | 12.769 | 17.371 |

\* Statistically significant at 99% CI

Table 3 shows that the change in speed is statistically significant at the 95% confidence interval post FCW by an average speed of 15.07 mph. To identify the most appropriate method to evaluate the factors influencing such a change in speed, three models were considered: a decision tree model, a random forest model, and an ordinary least squares regression model. The decision tree and the random forest models are machine learning models and they have a useful tool, called "variable importance," which ranks the variables according to their importance, as they relate to the dependent variable. To select the best model for this FCW dataset across machine learning and statistics, a comparison of R-squared values and Mean squared error (MSE) were considered to be the most appropriate. A higher R-squared value and a lower MSE would suggest the best fit model, out of the models considered for this dataset. The output of the comparison is shown in Table 4.

**Table 4.** Model Comparison

| Model | R-Squared | MSE |
| --- | --- | --- |
| Decision Trees | 0.268 | 101.3 |
| Random Forest | 0.577 | 65.4 |
| Linear Regression | 0.212 | 109.1 |

Based on Table 15, a Random Forest model with the highest R-squared value of 0.577 and the lowest MSE value of 65.4 was considered as the best fit for this dataset.

### 3.2. MDG Score – Driving Simulator

Figure 4 shows the MDG score for all the variables used for the change in speed analysis. It can be seen that "age" and "familiarity with CAVs" stand out and thus are selected as the most important variables that impact change in speed, post FCW. Figure 4 shows the variable importance scores for the respective variables, which means that the variables with the highest importance scores are the ones that give the best prediction and contribute the most to the model. This also means that if the top variables are dropped from the model, the predictive power of the model will be greatly reduced as compared to removing the least important variables.

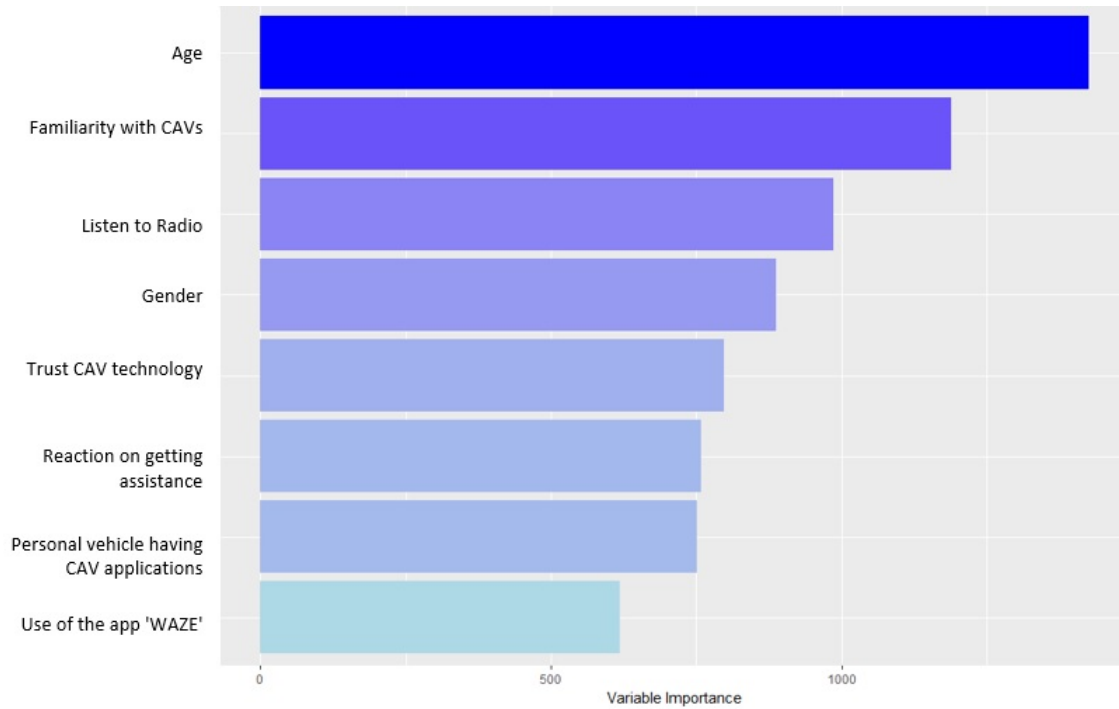

**Figure 4.** Variable importance based on increasing node impurity (Simulator)

Based on the descriptive statistics, with a change in speed between 15 and 30 mph, more than 66% of the participants were below the age of 35. Thus, it can be inferred that participants in the younger age group tend to slow down more when encountering an FCW, compared to the participants older than 35. Participants' familiarity with CAV technology could also positively or negatively affect speed change, post FCW.

### *3.3. Driving Simulator Data Validation*

The U.S. Department of Transportation, in collaboration with the University of Michigan Transportation Research Institute (UMTRI) and partner organizations, undertook a pilot called the Safety Pilot Model Deployment (SPMD) from August 2011 to February 2014 [38]. The objective of this pilot was to support the evaluation of Dedicated Short-Range Communications (DSRC) technology for V2V safety applications in the real world. The test area included three major east-west corridors in Ann Arbor, Michigan, where nearly 3000 equipped vehicles were deployed, on more than 70 miles of instrumented roadway, which makes it the largest connected vehicle technology field test in the world to date [38]. Data was collected on emergency electronic brake lights, forward collision warning, curve speed warning, intersection movement assist and basic safety warning messages. To evaluate the findings of this driving simulator study, all instances of FCW were obtained from UMTRI.

#### *3.3.1. Forward Collision Warning*

The data obtained from UMTRI consisted of 106 unique device recorders or participants, including 8,235 trips involving 12,210 instances of FCW. A radar unit installed in the vehicles, manufactured by Mobil Eye, could look ahead as far as 650 feet with a 60-degree field of view. The data parsed and shared for this study included average speeds 5 seconds before and after the FCW was issued.

Other variables provided were binary braking events, before and after the warning. The binary values signified whether braking had occurred before the FCW event and/or after the event. The data included information about turns or lane change indications, i.e., if the participants had their turn signal on before or after the FCW event, and whether they were left or right turn signals. The final variable used was daylight, i.e., if the trip took place during the daytime or at night.

*3.3.2. Data from Michigan Field Study*

The FCW data obtained from the Michigan study consisted of 106 unique vehicles/participants involved in 12,210 instances of FCW. To obtain metrics similar to the ones observed in the driving simulator, an average change in driving speed was calculated 5 seconds before and after the FCW was issued.

*3.4. One-sample t-test*

A one sample t-test was conducted to determine whether the mean difference in speed change is statistically different from the hypothesized mean difference in speed of zero.

**Table 5.** One sample t-test

| | | | Hypothesized Mean Difference = 0 | | | |
|---|---|---|---|---|---|---|
| | | | | | 95% Confidence Interval of the Difference | |
| | t | df | Sig. (2-tailed) | Mean Difference | Lower | Upper |
| Change in speed | 55.114 | 12209 | 0.000* | 8.597 | 8.291 | 8.903 |

\* Statistically significant at 95% CI

Table 5 shows that the change in speed is statistically significant at the 95% confidence interval post FCW by an average speed of 8.597 mph. Since sociodemographic variables could not be obtained for this dataset, the influence of the available variables, namely "braking before" and "braking after," could possibly bring some insight to the impact of FCW on speed change behavior. To identify the most appropriate method to evaluate these factors, again, three models were considered: a decision tree model, a random forest model, and an ordinary least squares regression model, in which a comparison of R-squared values and Mean squared error (MSE) were considered to be the most appropriate. The output of the comparison is shown in Table 6.

Table 6. Model Comparison

| Model | R-Squared | MSE |
|---|---|---|
| Decision Trees | 0.505 | 146.9 |
| Random Forest | 0.518 | 143.9 |
| Linear Regression | 0.477 | 155.1 |

Based on Table 17, a Random Forest model with the highest R-squared value of 0.518 and the lowest MSE value of 143.9 was considered as the best fit for this dataset.

### 3.5. MDG Score – Field Test

Figure 5 shows the MDG score for all the variables used for the change in speed analysis.

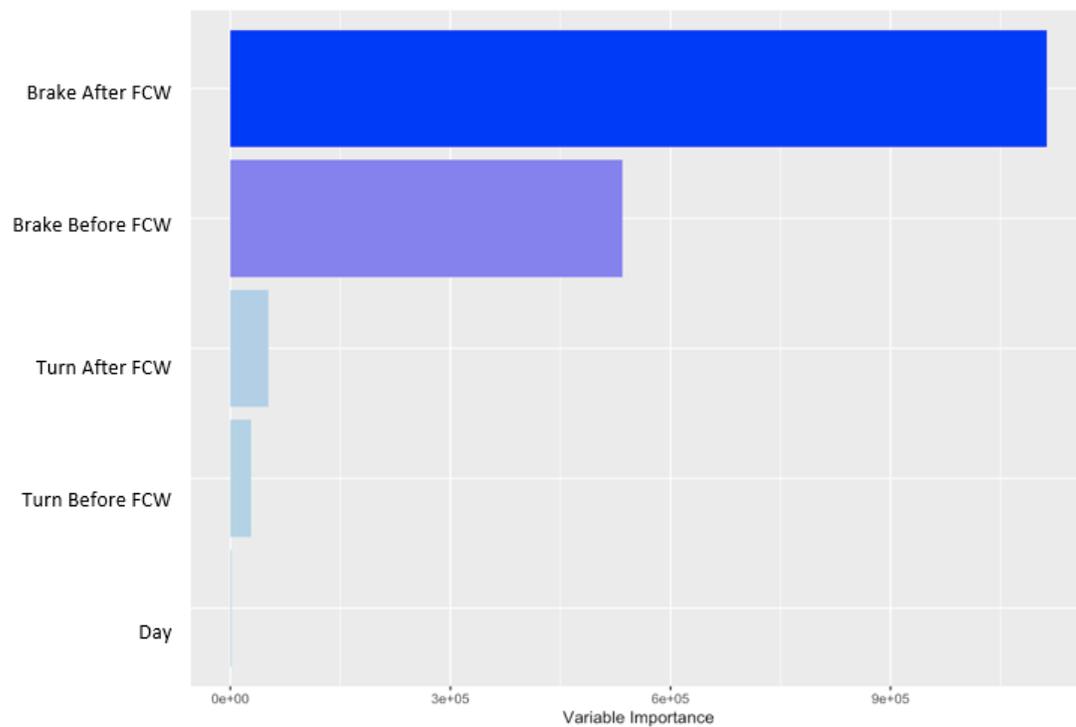

**Figure 5.** Variable importance based on increasing node impurity (Field Data)

It can be seen that braking after the FCW is the most important variable compared to the others. This means that the majority of the drivers only started braking after receiving the FCW compared to the drivers who had already started braking before receiving a warning.

### *3.6. Driving Simulator Validation*

If a two-sample t-test is performed to compare the changes in speed in both the driving simulator as well as the Michigan field data with regard to the effectiveness of FCW, the change in speeds of 15.07 mph and 8.597 mph will be significantly different. This is mainly due to the different speed limits of the roads in Ann Arbor, compared to the driving simulator. Comparing the mean difference in speeds is not the goal of this validation. This validation simply proves that an FCW system in a driving simulator is as effective as the ones used in the field to induce a speed reduction.

## 4. Conclusions and Discussion

This application, an FCW system, was tested on driver braking behavior using a full-scale, medium fidelity driving simulator and validated using field test data obtained from UMTRI. Ninety-three participants were involved in this study, consisting of 104 FCW instances. Results from the driving simulator study showed that, the FCW system had a statistically significant impact, at the 95% confidence interval, on the change in speed and the overall speed reduction of the vehicles, which was calculated at 5 seconds before and after the FCW was issued. This finding shows that these systems impact the drivers' performance positively. This conclusion matches the ones obtained by [16] and [26]. Moreover, our simulator experiment's findings proved that the familiarity with CAVs is an important factor that can impact the drivers' change of speed post FCW. This observation is in line with the one deduced by [22] who found that familiarity with warning systems has a significant impact on the drivers' performance. The final observation that can be deduced from our simulation experiment is related to the impact of the drivers' ages on the change in speed. Based on our descriptive statistics analysis, more than 66% of the participants who had a change in speed between 15 and 30 mph were below the age of 35; hence, it can be inferred that participants in the younger age group tend to slow down more when encountering an FCW, compared to participants older than 35. Nonetheless, this observation is at odds with most of the previous research studies on the impact of the FCW on the different drivers' behaviors. For instance, [18] found that the drivers' age did not impact their headway when an FCW is present, while [28] found that there was no significant difference between younger (below 45 years) and older drivers' (above 45 years) reaction times after receiving a warning from the FCW system. Finally, our real-world findings confirmed the conclusions drawn from the simulation experiment as it was also found that the presence of the FCW had a positive and significant impact on the change of speed, which proves the effectiveness of our simulation experiment. Moreover, this field experiment found that the majority of drivers started pressing the brake pedal after receiving the warning, which shows the effectiveness of the FCW system. Thus, from both the simulator and the field tests, it can be said that an FCW system is effective in inducing a speed reduction. Thus, simulator studies could be used as a standalone method to conducted behavioral studies related to connected and autonomous vehicle applications, in situations which do not permit field testing yet.

**Acknowledgments**

This study was supported by the Urban Mobility and Equity Center (Grant# 69A43551747123), a Tier 1 University Transportation Center of the U.S. DOT University Transportation Centers Program at Morgan State University. The authors would like to express their gratitude to Brian Lin at the University of Michigan Transportation Research Institute for sharing his valuable time and providing validation data for this study.

**Disclosure Statement**

No potential conflict of interest was reported by the authors.

**Funding**

This work was supported by the Urban Mobility and Equity Center, a Tier 1 University Transportation Center of the U.S. DOT University Transportation Centers Program at Morgan State University.

**References**

1. NTSB, *Addressing Deadly Rear-End Crashes*. 2015.
2. Institute, I.I. *Facts + Statistics: Highway safety*. 2019 [cited 2019 07/01/2019]; Available from: https://www.iii.org/fact-statistic/facts-statistics-highway-safety.
3. Bella, F. and R. Russo, *A collision warning system for rear-end collision: a driving simulator study.* Procedia-social and behavioral sciences, 2011. **20**: p. 676-686.
4. Lee, J.D., et al., *Driver distraction, warning algorithm parameters, and driver response to imminent rear-end collisions in a high-fidelity driving simulator*. 2002.
5. Kusano, K.D. and H.C. Gabler, *Safety benefits of forward collision warning, brake assist, and autonomous braking systems in rear-end collisions.* IEEE Transactions on Intelligent Transportation Systems, 2012. **13**(4): p. 1546-1555.
6. NHTSA, *Preliminary statement of policy concerning automated vehicles.* Washington, DC, 2013: p. 1-14.
7. Kockelman, K., et al., *An assessment of autonomous vehicles: traffic impacts and infrastructure needs*. 2017, University of Texas at Austin. Center for Transportation Research.
8. Patra, S., et al., *A Forward Collision Warning System for Smartphones Using Image Processing and V2V Communication.* Sensors, 2018. **18**(8): p. 2672.
9. Bham, G., et al., *Younger driver's evaluation of vehicle mounted attenuator markings in work zones using a driving simulator.* Transportation Letters, 2010. **2**(3): p. 187-198.
10. Shaw, F.A., et al., *Effects of roadway factors and demographic characteristics on drivers' perceived complexity of simulated roadway videos.* Transportation Letters, 2019. **11**(10): p. 589-598.
11. Banerjee, S., M. Jeihani, and N.K. Khadem, *Influence of work zone signage on driver speeding behavior.* Journal of Modern Transportation, 2019: p. 1-9.
12. Banerjee, S., et al., *Units of information on dynamic message signs: a speed pattern analysis.* European Transport Research Review, 2019. **11**(1): p. 15.
13. Banerjee, S., M. Jeihani, and R.Z. Moghaddam, *Impact of Mobile Work Zone Barriers on Driving Behavior on Arterial Roads.* Journal of Traffic and Logistics Engineering, 2018. **Vol. 6**(No. 2).
14. Banerjee, S., et al., *Influence of red-light violation warning systems on driver behavior–a driving simulator study.* Traffic injury prevention, 2020. **21**(4): p. 265-271.
15. Jeihani, M., et al., *Driver's Interactions with Advanced Vehicles in Various Traffic Mixes and Flows (Connected and Autonomous Vehicles (CAVs), Electric Vehicles (EVs), V2X, Trucks, Bicycles and Pedestrians)-Phase I: Driver Behavior Study and Parameters Estimation*. 2020.
16. Burns, P.C., E. Knabe, and M. Tevell. *Driver behavioral adaptation to collision warning and avoidance information*. in *Proceedings of the Human Factors and Ergonomics Society... Annual Meeting*. 2000. Sage Publications Ltd.
17. Lee, J.D., D.V. McGehee, and T.L. Brown. *Prior Exposure, Warning Algorithm Parameters and Driver Response to Imminent Rear-End Collisions on a High-Sfideltiy Simulator*. in *Proceedings of* *the human factors and ergonomics society annual meeting*. 2000. SAGE Publications Sage CA: Los Angeles, CA.
18. Shinar, D. and E. Schechtman, *Headway feedback improves intervehicular distance: A field study.* Human Factors, 2002. **44**(3): p. 474-481.
19. Maltz, M. and D. Shinar, *Imperfect in-vehicle collision avoidance warning systems can aid drivers.* Human factors, 2004. **46**(2): p. 357-366.
20. Lees, M.N. and J.D. Lee, *The influence of distraction and driving context on driver response to imperfect collision warning systems.* Ergonomics, 2007. **50**(8): p. 1264-1286.
21. Jamson, A.H., F.C. Lai, and O.M. Carsten, *Potential benefits of an adaptive forward collision warning system.* Transportation research part C: emerging technologies, 2008. **16**(4): p. 471-484.
22. Koustanaï, A., et al., *Simulator training with a forward collision warning system: Effects on driver-system interactions and driver trust.* Human factors, 2012. **54**(5): p. 709-721.
23. Muhrer, E., K. Reinprecht, and M. Vollrath, *Driving with a partially autonomous forward collision warning system: How do drivers react?* Human factors, 2012. **54**(5): p. 698-708.
24. Behbahani, H., N. Nadimi, and S. Naseralavi, *New time-based surrogate safety measure to assess crash risk in car-following scenarios.* Transportation Letters, 2015. **7**(4): p. 229-238.
25. Nadimi, N., D.R. Ragland, and A. Mohammadian Amiri, *An evaluation of time-to-collision as a surrogate safety measure and a proposal of a new method for its application in safety analysis.* Transportation Letters, 2019: p. 1-10.
26. Ben-Yaacov, A., M. Maltz, and D. Shinar, *Effects of an in-vehicle collision avoidance warning system on short-and long-term driving performance.* Human Factors, 2002. **44**(2): p. 335-342.
27. NHTSA, *Automotive Collision Avoidance System Field Operational Test*. 2005.
28. Crump, C., et al., *Driver Reactions in a Vehicle with Collision Warning and Mitigation Technology*. 2015, SAE Technical Paper.
29. FORUM8. *3D VR & Visual Interactive Simulation*. Available from: http://www.forum8.com/.
30. NACTO, *Vehicle Stopping Distance and Time*
31. Banerjee, S., M. Jeihani, and R. Moghaddam, *Impact of Mobile Work Zone Barriers on Driving Behavior on Arterial Roads.* Journal of Traffic and Logistics Engineering Vol, 2018. **6**(2).
32. Jeihani, M., et al., *The Potential Effects of Composition and Structure of Dynamic Message Sign Messages on Driver Behavior using a Driving Simulator*. 2018.
33. Moghaddam, Z.R., et al., *Comprehending the roles of traveler perception of travel time reliability on route choice behavior.* Travel Behaviour and Society, 2019. **16**: p. 13-22.
34. Breiman, L., *Random forests.* Machine learning, 2001. **45**(1): p. 5-32.
35. Breiman, L., *Bagging predictors.* Machine learning, 1996. **24**(2): p. 123-140.
36. Liaw, A. and M. Wiener, *Classification and regression by randomForest.* R news, 2002. **2**(3): p. 18-22.
37. Banerjee, S., M. Jeihani, and M.A. Sayed, *System and method for synchronization of asynchronous datasets*. 2021, Google Patents.
38. Bezzina, D. and J. Sayer, *Safety pilot model deployment: Test conductor team report.* Report No. DOT HS, 2014. **812**: p. 171.